\documentclass[10pt,twocolumn,letterpaper]{article}

\usepackage{cvpr}              % To produce the CAMERA-READY version
\usepackage{multirow}

\definecolor{cvprblue}{rgb}{0.21,0.49,0.74}
\usepackage[pagebackref,breaklinks,colorlinks,allcolors=cvprblue]{hyperref}

\def\paperID{32314} % *** Enter the Paper ID here
\def\confName{CVPR}
\def\confYear{2026}

\title{Reconstructing Spiking Neural Networks Using a Single Neuron with Autapses}

\author{
	Wuque Cai$^{1}$ \quad
	Hongze Sun$^{1}$ \quad
	Quan Tang$^{1}$ \quad
	Shifeng Mao$^{1}$ \quad
	Zhenxing Wang$^{1}$ \quad
	Jiayi He$^{1}$ \quad\\
	Duo Chen$^{2*}$ \quad
	Dezhong Yao$^{1*}$\quad
	Daqing Guo$^{1*}$\\
	$^{1}$Brain-Apparatus Communication Institute,\\ University of Electronic Science and Technology of China, Chengdu, China\\
	$^{2}$School of Artificial Intelligence, Chongqing University of Education, Chongqing, China\\
    {\tt\small $^{1}$\{dyao,dqguo\}@uestc.edu.cn, $^{2}$duochen3@gmail.com}
}

\begin{document}
\maketitle

\renewcommand{\thefootnote}{\fnsymbol{footnote}}
\footnotetext[1]{Corresponding authors.}

\begin{abstract}
Spiking neural networks (SNNs) are promising for neuromorphic computing, but high-performing models still rely on dense multilayer architectures with substantial communication and state-storage costs. Inspired by autapses, we propose time-delayed autapse SNN (TDA-SNN), a framework that reconstructs SNNs with a single leaky integrate-and-fire neuron and a prototype-learning-based training strategy. By reorganizing internal temporal states, TDA-SNN can realize reservoir, multilayer perceptron, and convolution-like spiking architectures within a unified framework. Experiments on sequential, event-based, and image benchmarks show competitive performance in reservoir and MLP settings, while convolutional results reveal a clear space--time trade-off. Compared with standard SNNs, TDA-SNN greatly reduces neuron count and state memory while increasing per-neuron information capacity, at the cost of additional temporal latency in extreme single-neuron settings. These findings highlight the potential of temporally multiplexed single-neuron models as compact computational units for brain-inspired computing.
\end{abstract}

\section{Introduction}
Spiking neural networks~(SNNs) are widely regarded as the third generation of neural networks~\cite{maass1997networks, ghosh2009spiking}. They serve as an important foundation for neuromorphic computing~\cite{roy2019towards}, featuring event-driven processing, high energy efficiency, and rich temporal dynamics~\cite{zhou2024direct, dampfhoffer2023backpropagation}. However, most high-performing SNNs still adopt dense multilayer architectures similar to artificial neural networks~(ANNs)~\cite{wu2018spatio,wu2019direct,hu2021spiking,sengupta2019going,wang2023masked}. Such designs require extensive neuron communication and large state storage, limiting spatial efficiency and hindering deployment in resource-constrained settings~\cite{zenke2015diverse,zhao2022framework,8267253}. This limitation motivates the search for compact SNN architectures that preserve computational capability while reducing structural redundancy.~\cite{yao2023inherent}

The intrinsic computational potential of single neurons provides a promising direction for advancing information processing. Biological neurons not only perform dendritic integration~\cite{spruston2008pyramidal, li2019dendritic} but also exhibit complex mechanisms such as intrinsic plasticity~\cite{debanne2019plasticity} and lateral inhibition~\cite{meinhardt2000pattern}, enabling rich nonlinear dynamics across spatiotemporal dimensions. In the SNN field, various strategies have been proposed to enhance single-neuron expressiveness, including dendritic computation~\cite{liu2024dendritic}, intrinsic plasticity~\cite{sun2023synapse, fang2021incorporating}, lateral inhibition–inspired competition mechanisms~\cite{zhang2021spiking, cheng2020lisnn}, neuronal heterogeneity~\cite{zheng2024temporal, perez2021neural}, and multi-compartment modeling~\cite{chen2024pmsn}. Although these approaches improve the nonlinear computational capability of individual neurons to some extent~\cite{li2024brain}, their internal states still rely mainly on instantaneous inputs, which limits their ability to retain long-term information or perform recursive computations~\cite{bikic2025cost}.
Among these, cerebellar Purkinje cells exemplify single-neuron complexity through elaborate dendrites and autaptic self-feedback, supporting temporal processing and memory.

In exploring the computational potential of single-neuron structures, existing engineering approaches based on single-neuron delay loops offer valuable insights. Studies on single-neuron deep learning~\cite{stelzer2021deep, stelzer2021emulating} and one-core reservoir computing~(RC) models~\cite{peng2025one} demonstrate that a single node possesses significant capacity for temporal information processing. However, these methods typically rely on non-biological hardware, and their continuous-value computations fundamentally differ from spike-based, event-driven mechanisms, limiting their direct applicability to SNNs and biological interpretation.

In biological neural systems, autapses provide a natural mechanism for temporal self-interaction. An autapse is a synaptic connection that a neuron forms with itself, enabling it to sense its own past spiking activity, and to exert either inhibitory or excitatory modulation on its current state~\cite{wang2017formation, baysal2023stochastic, yao2019inhibitory}. From a computational perspective, autapses introduce intrinsic temporal memory within neurons, naturally embedding the high-dimensional information into their dynamics and providing a biological basis for single neurons to realize long-term dependencies and recursive computation~\cite{seung2000autapse, yilmaz2016autapse}.
Such autaptic self-modulation in Purkinje cells inspires our model design for long-term dependencies and recursive computation.

Based on this observation, we propose a leaky integrate-and-fire neuron with delayed autapses~(TDA-LIF), which forms the basis of a corresponding spiking neural network model termed TDA-SNN. Under different structural configurations, the TDA-LIF neuron reorganizes its internal temporal-node evolution to construct reservoir-computing, multilayer-perceptron, and convolution-like spiking architectures within a unified framework. Experiments on sequential, event-based, and image datasets show that TDA-SNN provides competitive performance in reservoir and MLP settings, while its convolutional results reveal a clear space--time trade-off. Compared with structurally equivalent standard SNNs, TDA-SNN drastically reduces neuron count and state memory and increases per-neuron information capacity, at the cost of additional temporal latency in extreme single-neuron settings. Our main contributions are summarized as follows:
\begin{itemize}
	\item We theoretically show that a TDA-LIF neuron can constructively realize three representative SNN structures: reservoir computing~(RC), multilayer perceptrons~(MLPs), and convolution-like architectures.
	\item We propose a dedicated optimization framework for TDA-SNN to address the training challenges introduced by internal delayed feedback, enabling stable learning of temporally unfolded single-neuron models.
	\item Extensive experiments on multiple benchmarks demonstrate that TDA-SNN achieves competitive performance in RC and MLP settings, and reveal a clear space--time trade-off in convolutional settings, while substantially reducing neuron count and state memory and significantly enhancing per-neuron information capacity.
\end{itemize}

\section{Related Works}
\subsection{Neurons with High Computing Capacity}
Early artificial neuron models, such as McCulloch-Pitts neurons~\cite{mcculloch1943logical}, greatly simplify the richer dynamics of biological neurons, whose dendrites, ion channels, and nonlinear membrane properties support powerful local computation.~\cite{calimera2013human} In particular, dendrites can integrate inputs, perform nonlinear operations, and even implement logical functions~\cite{spruston2008pyramidal,li2019dendritic,gidon2020dendritic}, motivating SNN studies that enhance single-neuron capability through dendritic computation, intrinsic plasticity, lateral inhibition, neuronal heterogeneity, and multi-compartment modeling.~\cite{liu2024dendritic,sun2023synapse,fang2021incorporating,zhang2021spiking,cheng2020lisnn,zheng2024temporal,perez2021neural,chen2024pmsn,li2024brain} However, the role of temporal delay in improving single-neuron computation remains underexplored.~\cite{beniaguev2021single,cavanagh2020diversity,bikic2025cost}

\subsection{Time-Delay Neural Network}
Transmission delay is a fundamental resource for temporal processing in both biological and artificial neural systems.~\cite{bialek1992reliability,shavikloo2024synchronization} Time-delay neural networks explicitly model temporal dependencies through delayed connections~\cite{waibel2013phoneme,ji2024trainable}, and folded-in-time deep networks further compress depth into delay loops within a single node, achieving strong performance in vision and EEG tasks.~\cite{stelzer2021deep,stelzer2021emulating,peng2025one} Yet these approaches typically rely on continuous-valued signals and have limited biological plausibility. In contrast, biological autapses allow neurons to regulate current activity using their own past spikes.~\cite{wang2017formation,baysal2023stochastic,yao2019inhibitory} This perspective motivates us to view delay as an intrinsic autaptic property. It enables a single spiking neuron to emulate network-like temporal computation while preserving interpretability and improving long-range temporal modeling.

%%%%%%%%%%%%%%%%%%%%%%%%%%%%%%%%%%%%%%%%%%%%%%%%Figure1
\begin{figure}[t]
	\centering
	\includegraphics{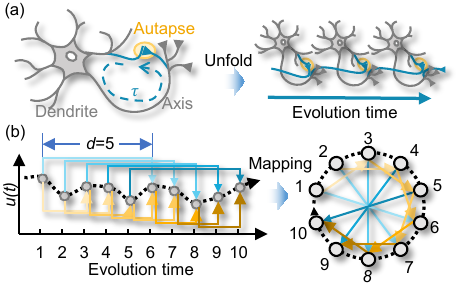}
	\caption{Evolution and folding principles of autapses. (a) An autapse in biological neurons and its evolutionary unfolding. (b) Distribution of evolutionary time in the TDA-LIF model and its mapping to reservoir computation.}
	\label{Fig1}
\end{figure}
%%%%%%%%%%%%%%%%%%%%%%%%%%%%%%%%%%%%%%%%%%%%%%%%

\section{Method}
We introduce time-delayed autapses into the LIF neuron to enhance its temporal dynamics and internal memory. Based on this modification, we construct a temporally unfolded single-neuron model and show how it can be mapped to reservoir-computing, MLP, and convolution-like spiking architectures. Unless otherwise stated, $t$ denotes the index of an internal temporal node, while $T$ denotes the external data time window used for sequential or event-based inputs.
%%%%%%%%%%%%%%%%%%%%%%%%%%%%%%%%%%%%%%%%%%%%%%%%Leak Integrated-and-Fire Neuron with Time-delayed Autapses
%%%%%%%%%%%%%%%%%%%%%%%%%%%%%%%%%%%%%%%%%%%%%%%%%Figure2
%\begin{figure}[t]
%	\centering
%	\includegraphics{fig2}
%	\caption{Reservoir computing of TDA-LIF model.}
%	\label{Fig2}
%\end{figure}
%%%%%%%%%%%%%%%%%%%%%%%%%%%%%%%%%%%%%%%%%%%%%%%%%

\subsection{LIF Neuron with Time-delayed Autapses}
% In the study of SNNs, various neuron models have been employed, including the Hodgkin–Huxley model~\cite{hodgkin1952quantitative}, the Izhikevich model~\cite{izhikevich2003simple}, and the leaky integrate-and-fire~(LIF) model~\cite{gerstner2002spiking}. Among these, the LIF model has become the most widely used, as it provides a favorable trade-off between biological plausibility and computational efficiency. Because of this study emphasizes the design of connection structures, we adopt the LIF neuron model as the fundamental unit. The discretized iteration formula for the LIF model is as follows:
% \begin{equation} 
% 	v^{t} = \tau v^{t-1}(1-s^{t-1}) + I^t, \label{eq1}
% \end{equation}
% where $\tau$ denotes the membrane decay constant of the neuron, $v^{t-1} \in \mathbb{R}$ and $s^{t-1} \in {0,1}$ represent the membrane potential and the spike of the neuron at timestep $t-1$, respectively, and $I^t = W s^t$ denotes the integrated input current from the presynaptic neurons at timestep $t$, with $W$ being the synaptic weight matrix. Once the membrane potential exceeds the threshold, a spike is output. This process is described as
% \begin{align}
% 	s^{t} = H(v^{t}) = \begin{cases} \displaystyle 1, & \text {if}~v^{t}\ge v_{\text {th}}\\ \displaystyle 0, & \text{otherwise} \end{cases},
% 	\label{eq2}
% \end{align}
% where $H(\cdot)$ is the Heaviside step function, which is non-differentiable at the step point.

Introducing an autapse with delay $d$ into the LIF neuron~\cite{gerstner2002spiking} provides temporal self-feedback: a spike emitted by the neuron is fed back to its dendrites after $d$ internal nodes. As shown in Fig.~\ref{Fig1}, the axon connects to the neuron's own dendrites, allowing past spikes to influence future membrane-potential evolution.

Mathematically, the iterative formula for the TDA-LIF is as follows:
\begin{equation} 
	v^{t} = \tau v^{t-1}(1-s^{t-1}) + I^t + I_{autapse}^t, \label{eq3}
\end{equation}
where $I_{autapse}^t = \sum_{d \in D_t} w_a^d s^{t-d}$ is the delayed autaptic current generated by the neuron's own past spikes, $D_t=\{d\in\mathbb{N}\mid d<t\}$ is the set of valid delays at node $t$, and $w_a^d$ denotes the weight of the autapse with delay $d$.

%%%%%%%%%%%%%%%%%%%%%%%%%%%%%%%%%%%%%%%%%%%%%%%%Temporal Dynamics to Reservoir Computing
\subsection{Temporal Dynamics to Reservoir Computing}
Based on the TDA-LIF model, we unfold neuron dynamics into a sequence of temporal nodes, where autapses connect nodes across different internal timesteps. We refer to each internal timestep as a node to distinguish it from the external data time window. These connections are directional and transmit signals forward along the unfolded trajectory. For simplicity, autapses with the same delay are treated as one delay type. As shown in Fig.~\ref{Fig1}, the neuron evolves over ten nodes, with blue and yellow autapses representing delays of five and two nodes, respectively.

When autapses with different delays are assigned different weights, delayed self-feedback modulates the membrane-potential trajectory and enriches the internal temporal state. Under a finite set of delays, the membrane dynamics can be written as:
\begin{equation} 
	v^{t} = \tau v^{t-1}(1-s^{t-1}) + Wx^t + \sum_{d \in D} w^d_a(t)s^{t-d}, \label{eq4}
\end{equation}
where $x^t$ is the external input injected into node $t$, $D$ denotes the set of selected autaptic delays, $w^d_a(t)$ is the weight of the autapse with delay $d$ at node $t$, and $W \in \mathbb{R}^{N \times N_\text{in}}$ represents the weight matrix for the external inputs, with $N$ being the number of nodes and $N_\text{in}$ the input dimension.

When arranged as in the right panel of Fig.~\ref{Fig1}(b), the unfolded temporal graph induces an RC-like connectivity pattern. Signals propagate along the nodes following the dotted arrows, while delayed autaptic feedback continuously modulates the hidden dynamics. The resulting interaction across temporal nodes enriches the internal states, and aggregating neuronal responses over the unfolded trajectory yields a high-dimensional representation for reservoir computing.
 
%Based on this formulation, external inputs are fed into each node, and the firing activities of all neurons are recorded over a time window $T$. By comparing these firing patterns (prototypes), the input data can be effectively identified. Correspondingly, the autosynaptic connection matrix $W_a$, represents the temporal feedback structure among neuron time nodes, where each element denotes an autosynaptic connection from one time node to another. Owing to the directed nature of temporal feedback, $W_a$ takes the form of an upper triangular matrix, with each parallel diagonal corresponding to a specific delay in the autapse. This overall structure, unfolded from TDA-LIF along the temporal nodes and functionally equivalent to it, constitutes the complete SNN model referred to as TDA-SNN.

%%%%%%%%%%%%%%%%%%%%%%%%%%%%%%%%%%%%%%%%%%%%%%%%Figure2
\begin{figure}[t]
	\centering
	\includegraphics{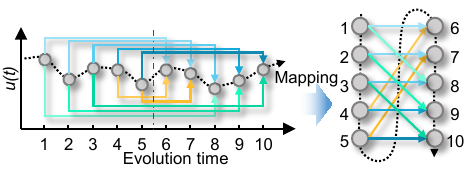}
	\caption{Division of the evolutionary time in the TDA-LIF model and its mapping to an MLP}
	\label{Fig2}
\end{figure}
%%%%%%%%%%%%%%%%%%%%%%%%%%%%%%%%%%%%%%%%%%%%%%%%

%%%%%%%%%%%%%%%%%%%%%%%%%%%%%%%%%%%%%%%%%%%%%%%%MLP Equivalence by Autapse Pruning
\subsection{MLP Equivalence by Autapse Pruning}
As shown in Fig.~\ref{Fig2}, the unfolded trajectory is divided into two segments, highlighted by the red dashed line, and autapses are pruned accordingly. Specifically, autapses confined within the same segment are removed, while those crossing the segment boundary are retained. Rearranging the temporal nodes into two groups then yields a feedforward network structure.

This construction produces an MLP-like structure from a single TDA-LIF neuron. In Fig.~\ref{Fig2}, the neuron integrates three delayed autaptic groups, $d = 2$ (yellow), $5$ (blue), and $7$ (green), yielding ten effective synaptic connections between the two temporal segments. During the evolution of the second segment, the dynamics can be written as:
\begin{equation} 
	v^{t} = \tau v^{t-1}(1-s^{t-1}) + Wx^t + \sum_{d \in D_{\text{FC}}} w^d_a(t)s^{t-d}, \label{eq5}
\end{equation}
where $D_{\text{FC}}=\{ d \in \mathbb{N} \mid d \le t,\, t-d < t_s\}$ denotes the set of delays that connect the first segment to the second, and $t_s$ is the segment boundary.

Building on this foundation, the TDA-LIF model can be extended to deeper feedforward networks by introducing more temporal segments. Inputs are assigned to the first segment, and spike responses from the final segment are used for recognition. In the corresponding inter-segment connection matrix, the diagonal and parallel-diagonal elements represent autaptic connections with identical delays.

%%%%%%%%%%%%%%%%%%%%%%%%%%%%%%%%%%%%%%%%%%%%%%%%Figure3
\begin{figure}[t]
	\centering
	\includegraphics{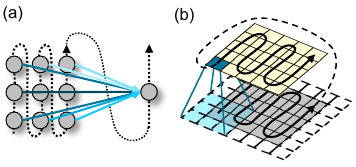}
	\caption{Convolutional layer with the TDA-LIF model. (a) Mapping 10 nodes to a 3×3 convolution. (b) Equivalence of TDA-LIF to a single-channel convolutional layer.}
	\label{Fig3}
\end{figure}
%%%%%%%%%%%%%%%%%%%%%%%%%%%%%%%%%%%%%%%%%%%%%%%%

%%%%%%%%%%%%%%%%%%%%%%%%%%%%%%%%%%%%%%%%%%%%%%%%Convolutional Structure Equivalence
\subsection{Convolutional Structure Equivalence}
%The core strength of convolutional neural networks (CNNs) lies in their ability to achieve translation-invariant feature extraction across spatial dimensions through weight sharing and local connectivity. This section demonstrates that a single TDA-LIF neuron can equivalently implement the convolution operation within its dynamical framework by introducing a specific spatiotemporal connectivity configuration.

Starting from the MLP-like topology, we assign each spatial output node a set of autapses with shared weights and corresponding delays, thereby encoding the weight-sharing mechanism of a convolution kernel into the neuron's internal temporal structure. These shared autaptic weights play the role of kernel parameters, while distinct delay groups provide a temporal realization of local spatial aggregation.

During signal propagation, output nodes operate in parallel and synchronously, each using the shared autaptic kernel to process a different local region of the input. In this way, kernel translation across space is mapped to a repeated delay-sharing pattern across temporal nodes. As illustrated in Fig.~\ref{Fig3}, this construction provides a convolution-like realization in the spike domain and unifies local spatial aggregation with temporal dynamics.

For example, consider a convolution with a $3\times 3$ kernel, padding 1, and stride 1, with a single input and output channel for simplicity. The TDA-LIF state at a given node is determined by delayed inputs from nine neighboring positions in the previous layer, corresponding to nine delay groups. Because all spatial locations share the same delayed autaptic weights, the unfolded dynamics recover the key properties of local connectivity and weight sharing. This yields a constructive convolution-like mapping, which we later evaluate empirically in a preliminary setting.

%%%%%%%%%%%%%%%%%%%%%%%%%%%%%%%%%%%%%%%%%%%%%%%%Figure4
\begin{figure}[t]
	\centering
	\includegraphics{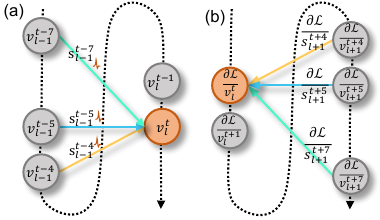}
	\caption{Forward (a) and backward (b) processes in the TDA-SNN model.}
	\label{Fig4}
\end{figure}
%%%%%%%%%%%%%%%%%%%%%%%%%%%%%%%%%%%%%%%%%%%%%%%%

%%%%%%%%%%%%%%%%%%%%%%%%%%%%%%%%%%%%%%%%%%%%%%%%Prototype Learning Method
\subsection{Prototype Learning Method}
To effectively decode neuronal firing patterns, it is essential to exploit their intrinsic spatio-temporal dynamics. Conventional static decoding methods neglect temporal depen dencies and fail to align with the spiking nature of SNN outputs. To address this limitation, we introduce a spatiotemporal prototype learning framework that matches output spike sequences to learnable prototypes, enabling efficient and interpretable recognition of dynamic firing behaviors~\cite{cai2025robust}. See the supplementary materials for additional details.
To effectively decode neuronal firing patterns, it is essential to exploit their intrinsic spatio-temporal dynamics. Conventional static decoding methods neglect temporal dependencies and fail to align with the spiking nature of SNN outputs. To address this limitation, we introduce a spatiotemporal prototype-learning framework that matches output spike sequences to learnable prototypes, enabling efficient and interpretable recognition of dynamic firing behaviors~\cite{cai2025robust}. Additional implementation details are provided in the supplementary material.

Specifically, a set of binary prototypes $\mathbf{K} \in \{0, 1\}^{C \times T}$ is defined to represent target spiking patterns for $C$ classes over $T$ steps. Let $f(x;\theta)$ denote the encoded spatiotemporal representation of input $x$. The similarity between $f(x;\theta)$ and the $i$-th prototype $\mathbf{k}_i \in \mathbf{K}$ is measured by
\[
	d_i(x) = -\left\|f(x;\theta)-\mathbf{k}_i\right\|_2^2.
\]

Based on the distances between the encoded representation and class prototypes, the classification objective is formulated as:
\begin{equation} 
	\mathcal{L} = -\log{\frac{e^{d_i\left(x\right)}}{\sum_{j=1}^C{e^{d_j\left(x\right)}}}} - \lambda d_i(x), \label{eq6}
\end{equation}
where $C$ represents the number of classes, and $\lambda$ is a hyperparameter controlling the contribution of the prototype regularization term (set to 0.001 in our experiments).

For network optimization, we adopt the spatio-temporal backpropagation (STBP) algorithm, which has been widely used in SNNs due to its efficiency in handling temporal dependencies.~\cite{wu2018spatio,wu2019direct} To overcome the non-differentiability of spike firing in TDA-LIF neurons and prototype binarization, we replace the gradient with a smooth approximation using the arctangent function:
\begin{equation} 
	s^{t} = g\left ({v^t}\right) \approx \frac{1}{\pi}\text{arctan}\left[\frac{\pi}{2}\alpha (v^{t}-v_{\text{th}})\right] + \frac{1}{2}. \label{eq7}
\end{equation}

Fig.~\ref{Fig4} illustrates information flow in TDA-SNN. In forward propagation, node information arises from both delayed autaptic spikes and the membrane potential carried over from the previous node, as shown in Fig.~\ref{Fig4}(a). Accordingly, in backward propagation, the gradient at each node is determined by the membrane-potential gradients of its successor node and of the future nodes that receive its delayed spikes, as shown in Fig.~\ref{Fig4}(b).

For the RC and MLP settings, the membrane-potential gradient at node $t$ is:
\begin{equation}
	\begin{aligned}
		\nabla_{v^t} \mathcal{L} & = \nabla_{v^{t+1}} \mathcal{L} \cdot \frac{\partial v^{t+1}}{\partial v^{t}} + \sum_{d \in D} \left [ \nabla_{v^{t+d}} \mathcal{L} \cdot \frac{\partial v^{t+d}}{\partial v^{t}} \right ]   \\
		                         & = \nabla_{v^{t+1}} \mathcal{L} \cdot \tau (1-s^{t-1}) \\
		                         &\qquad+ \sum_{d \in D} \left [ \nabla_{v^{t+d}} \mathcal{L} \cdot w^d_a \cdot \frac{\partial s^{t}}{\partial v^{t}} \right ]
	\end{aligned},
	\label{eq8}
\end{equation}
where $w^d_a$ is the autaptic weight connecting the spike at node $t$ to the future node $t+d$. The corresponding weight gradients are
\begin{equation}
	\begin{aligned}
		\nabla_{w} \mathcal{L} & = \nabla_{v^t} \mathcal{L} \cdot x^t, \\
		\nabla_{w_a^d } \mathcal{L} & = \nabla_{v^{t+d}} \mathcal{L} \cdot s^{t}.
	\end{aligned}
	\label{eq9}
\end{equation}

Using equations (\ref{eq6})--(\ref{eq9}), TDA-SNN can be trained end-to-end with surrogate gradients. %The overall training process is summarized in Algorithm~\ref{arg1}.  

\section{Experiments}
%In this section, we evaluate the effectiveness of the proposed TDA-SNN across multiple benchmark datasets. First, we describe the experimental setup, including the datasets used and specific implementation details. Subsequently, we experimentally explore and analyze the method's performance across various network architectures. Finally, we validate the method's efficiency through computational complexity analysis.

%%%%%%%%%%%%%%%%%%%%%%%%%%%%%%%%%%%%%%%%%%%%%%%%Dataset and Implementation Details
\subsection{Experimental Setup}

We evaluate TDA-SNN across multiple benchmarks under different structural mappings. DEAP~\cite{koelstra2011deap} and SHD~\cite{cramer2020heidelberg} are used to evaluate the reservoir-computing setting; MNIST~\cite{deng2012mnist}, Fashion-MNIST~(fMNIST)~\cite{xiao2017fashion}, and DVS Gesture~\cite{amir2017low} are used to validate the MLP setting; and DVS Gesture together with CIFAR10~\cite{krizhevsky2009learning} are further used to study the convolution-like setting. All models are implemented in PyTorch and trained on an NVIDIA A100 GPU using the Adam optimizer with a cosine learning-rate decay schedule. Unless otherwise stated, training is conducted for 100 epochs. For RC and MLP experiments, results are averaged over 10 independent runs and reported as mean $\pm$ standard deviation; due to the higher computational cost of convolutional experiments, those results are averaged over 5 runs. To ensure a fair comparison, TDA-SNN and the corresponding standard SNN baselines use aligned training protocols and matched architectural scales for each setting. The complete hyperparameter configurations are provided in Supplementary Sec.~\ref{sec9}.

\begin{table*}
	\caption{Experimental Results of Autaptic Selection Strategies Across Different Datasets}
	\label{tab:rc_and_mlp}
	\centering
	\begin{tabular}{@{}ccccccccc@{}}
		\toprule
		       Dataset         & Delay  &        1        &            2            &       4        &       8        &           16            &           32            &           64            \\ \midrule
		 \multirow{2}*{DEAP}   &   MC   & 83.52$\pm$0.52  &     83.10$\pm$1.10      & 83.63$\pm$0.73 & 83.91$\pm$1.14 &     83.85$\pm$0.59      & \textbf{84.66$\pm$0.62} &     84.47$\pm$0.99      \\
		                       &   RD   & 83.62$\pm$1.12  &     83.34$\pm$1.14      & 83.48$\pm$0.64 & 83.73$\pm$1.02 &     83.63$\pm$0.76      &     83.98$\pm$0.81      & \textbf{84.47$\pm$0.99} \\ \midrule
		  \multirow{2}*{SHD}   &   MC   & 78.35$\pm$0.61  &     72.09$\pm$1.17      & 73.59$\pm$0.67 & 74.69$\pm$1.28 &     77.07$\pm$0.48      &     79.36$\pm$0.31      & \textbf{80.42$\pm$0.39} \\
		                       &   RD   & 78.55$\pm$0.68  &     78.62$\pm$0.46      & 77.60$\pm$2.49 & 78.58$\pm$0.25 &     77.63$\pm$1.87      &     79.26$\pm$1.21      & \textbf{80.42$\pm$0.39} \\ \midrule
		 \multirow{3}*{MNIST}  &   MC   & 33.53$\pm$3.51  & \textbf{94.44$\pm$0.10} & 94.39$\pm$0.16 & 94.33$\pm$0.13 &     94.37$\pm$0.15      &     94.42$\pm$0.20      &     94.38$\pm$0.12      \\
		                       &   RD   & 84.03$\pm$19.14 &     93.32$\pm$1.19      & 94.82$\pm$0.15 & 95.11$\pm$0.21 &     95.63$\pm$0.14      &     95.98$\pm$0.11      & \textbf{96.42$\pm$0.09} \\
		                       & T Inv. & 90.33$\pm$4.94  &     93.21$\pm$1.84      & 93.93$\pm$0.47 & 94.37$\pm$0.24 &     94.37$\pm$0.20      &     94.42$\pm$0.17      & \textbf{94.72$\pm$0.14} \\ \midrule
		\multirow{3}*{fMNIST}  &   MC   & 34.25$\pm$2.78  & \textbf{86.63$\pm$0.11} & 86.59$\pm$0.16 & 86.60$\pm$0.25 &     86.58$\pm$0.18      &     86.63$\pm$0.15      &     86.55$\pm$0.17      \\
		                       &   RD   & 80.51$\pm$14.05 &     85.68$\pm$1.28      & 86.54$\pm$0.22 & 86.88$\pm$0.14 &     87.23$\pm$0.13      &     87.40$\pm$0.23      & \textbf{87.55$\pm$0.16} \\
		                       & T Inv. & 83.62$\pm$3.30  &     85.90$\pm$0.86      & 86.32$\pm$0.24 & 86.53$\pm$0.21 &     86.62$\pm$0.20      &     86.59$\pm$0.16      & \textbf{86.74$\pm$0.19} \\ \midrule
		  \multirow{2}*{DVS}   &   MC   & 23.13$\pm$2.86  &     52.88$\pm$1.82      & 52.71$\pm$1.67 & 52.88$\pm$2.32 & \textbf{55.28$\pm$2.12} &     54.44$\pm$3.10      &     53.19$\pm$2.64      \\
		\multirow{2}*{Gesture} &   RD   & 45.45$\pm$8.79  &     51.84$\pm$5.32      & 53.61$\pm$1.91 & 55.31$\pm$2.92 &     57.88$\pm$1.94      &     58.40$\pm$1.69      & \textbf{59.55$\pm$2.04} \\
		                       & T Inv. & 40.87$\pm$8.15  &     50.21$\pm$6.71      & 54.34$\pm$2.02 & 54.79$\pm$3.31 &     55.38$\pm$2.35      &     55.42$\pm$2.68      & \textbf{57.01$\pm$2.48} \\ \bottomrule
	\end{tabular}
\end{table*}

%%%%%%%%%%%%%%%%%%%%%%%%%%%%%%%%%%%%%%%%%%%%%%%%Reservoir Computing Equivalence
\subsection{Reservoir Computing Equivalence}
To assess the RC capability of TDA-SNN, we conducted experiments on the DEAP and SHD datasets. We further investigate how reservoir size and autaptic selection strategies affect connectivity patterns and classification performance.

%In the RC architecture, the TDA-SNN connection matrix is an upper-triangular square matrix. Fig.~\ref{Fig6}(a) illustrates how the number of nodes and the autaptic selection strategy shape this matrix. Fewer nodes reduce the number of possible connections (middle matrix). Meanwhile, selecting autapses from the main diagonal to the upper-right corner using the maximum connections (MC) strategy maximizes connectivity while simultaneously minimizing latency (right matrix). In contrast, the random delay (RD) strategy produces sparser but longer-delay connections (left matrix).

%%%%%%%%%%%%%%%%%%%%%%%%%%%%%%%%%%%%%%%%%%%%%%%%Figure5
\begin{figure}[t]
	\centering
	\includegraphics{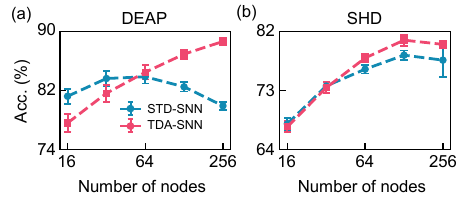}
	\caption{Comparison of STD-SNN and TDA-SNN performance under the RC structure with different reservoir sizes on (a) DEAP and (b) SHD datasets.}
	\label{Fig5}
\end{figure}
%%%%%%%%%%%%%%%%%%%%%%%%%%%%%%%%%%%%%%%%%%%%%%%%

%%%%%%%%%%%%%%%%%%%%%%%%%%%%%%%%%%%%%%%%%%%%%%%%Analysis of RC Equivalence
\subsubsection{Analysis of RC Equivalence}
To compare the performance of the standard SNN~(STD-SNN) and the proposed TDA-SNN with varying reservoir sizes, we conducted a series of experiments. Specifically, the number of reservoir neurons in the SNN was matched to the number of internal evolution nodes in the TDA-SNN to ensure comparable model complexity, thereby enabling a fair assessment of their representational and learning capabilities. As the reservoir size increased, the projection space became richer and more structured, leading to improved feature separability and discriminative performance. In this ablation study, reservoir sizes of 16, 32, 64, 128, and 256 were evaluated, with fully delayed autapses employed in all configurations.

As shown in Figs.~\ref{Fig5}(a) and (b), TDA-SNN underperforms STD-SNN at small reservoir sizes (e.g., 16 nodes), reaching 77.59$\pm$1.24\% versus 81.21$\pm$1.00\% on DEAP and 67.45$\pm$0.84\% versus 67.92$\pm$0.92\% on SHD. As the reservoir size increases, TDA-SNN improves steadily and becomes competitive with, or even surpasses, STD-SNN at larger sizes. At 256 nodes, TDA-SNN reaches 88.65$\pm$0.48\% on DEAP and 80.04$\pm$0.51\% on SHD, while STD-SNN attains 79.92$\pm$0.54\% and 77.63$\pm$2.53\%, respectively. On SHD, the best TDA-SNN result is obtained at 128 nodes (80.68$\pm$0.83\%), suggesting that excessive internal evolution can eventually introduce redundant temporal encoding. Overall, these results indicate that delayed autaptic dynamics become increasingly effective as the unfolded reservoir grows, making TDA-SNN a competitive alternative to RC-style SNNs in medium- and large-scale settings.
As shown in Figs.~\ref{Fig5}(a) and (b), TDA-SNN underperforms STD-SNN at small reservoir sizes (e.g., 16 nodes), reaching 77.59$\pm$1.24\% versus 81.21$\pm$1.00\% on DEAP and 67.45$\pm$0.84\% versus 67.92$\pm$0.92\% on SHD. As the reservoir size increases, TDA-SNN improves steadily and becomes competitive with, or even surpasses, STD-SNN at larger sizes. At 256 nodes, TDA-SNN reaches 88.65$\pm$0.48\% on DEAP and 80.04$\pm$0.51\% on SHD, while STD-SNN attains 79.92$\pm$0.54\% and 77.63$\pm$2.53\%, respectively. On SHD, the best TDA-SNN result is obtained at 128 nodes (80.68$\pm$0.83\%), suggesting that excessive internal evolution can eventually introduce redundant temporal encoding. These results indicate that delayed autaptic dynamics become increasingly effective as the unfolded reservoir grows, making TDA-SNN a competitive alternative to RC-style SNNs in medium- and large-scale settings.

%%%%%%%%%%%%%%%%%%%%%%%%%%%%%%%%%%%%%%%%%%%%%%%%Effect of Strategy Selection
\subsubsection{Effect of autaptic selection strategy}
This subsection investigates the influence of autaptic selection strategies on the performance of TDA-SNN under the RC framework. The reservoir size was fixed at 128 neurons, and two strategies, random delay (RD) and maximum connection (MC), were compared across different numbers of delayed autapses (1, 2, 4, 8, 16, 32, and 64). 

Table~\ref{tab:rc_and_mlp} summarizes the classification performance of TDA-SNN under different autaptic selection strategies on DEAP and SHD. On DEAP, both strategies benefit from increasing the number of delayed autapses, and MC shows a slight advantage at larger delay counts (32 and 64), indicating that denser structured delay assignment can improve information propagation when the task is relatively stable. On SHD, the trend is less monotonic: MC drops markedly at two delays, then recovers as more delayed connections are introduced. This observation suggests that on temporally richer datasets, overly constrained delay patterns may be suboptimal in low-delay regimes, whereas sufficient delayed connections allow structured feedback to recover its representational advantage. Thus, the ablation indicates that both the number of autapses and the delay-selection strategy materially affect RC performance, and that moderate-to-large delay sets are generally more reliable than sparse configurations.
Statistical significance analyses of the experimental results are provided in Supplementary Sec.~\ref{sec11}.

%%%%%%%%%%%%%%%%%%%%%%%%%%%%%%%%%%%%%%%%%%%%%%%%Multilayer Perceptron Equivalence
\subsection{Multilayer Perceptron Equivalence}
To further evaluate the generality of TDA-SNN in feedforward settings, we extended the experiments to MLP-like architectures. Specifically, we investigated whether the time-delayed autaptic mechanism can reproduce feedforward spiking representations through temporal unfolding. Following the RC analysis, we examined the effects of temporal-node number, autaptic selection strategy, and network depth on MNIST, fMNIST, and DVS Gesture.

%%%%%%%%%%%%%%%%%%%%%%%%%%%%%%%%%%%%%%%%%%%%%%%%Figure6
\begin{figure}[t]
	\centering
	\includegraphics{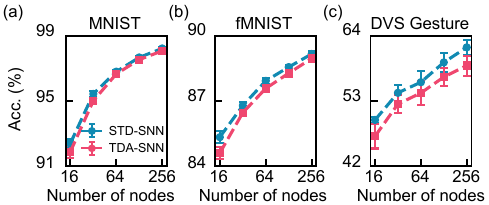}
	\caption{Comparison of TDA-SNN and STD-SNN accuracy under MLP architectures with varying numbers of nodes on (a) MNIST, (b) fMNIST, and (c) DVS Gesture.}
	\label{Fig6}
\end{figure}
%%%%%%%%%%%%%%%%%%%%%%%%%%%%%%%%%%%%%%%%%%%%%%%%

%For the MLP architecture, the TDA-SNN connection matrix width and height correspond to the number of input and output nodes, respectively. Fig.~7(a) illustrates how self-synapse selection strategies shape the matrix. Using the maximum connectivity (MC) strategy, self-synapses are selected along the diagonal and distributed to neighboring nodes, maximizing connectivity while minimizing latency. In contrast, the random delay (RD) strategy yields sparser, longer-delayed connections. This demonstrates that, in feedforward structures, structured delay assignment enables efficient temporal propagation across layers, analogous to the role of weight connectivity in conventional MLPs.

%%%%%%%%%%%%%%%%%%%%%%%%%%%%%%%%%%%%%%%%%%%%%%%%MLP Structural Equivalence
\subsubsection{MLP Structural Equivalence}
To validate the structural equivalence and representational capability of TDA-SNNs in multilayer feedforward architectures, we compared their performance with STD-SNNs under identical MLP configurations. The network comprises two layers: the first layer performs preliminary feature extraction from external inputs, while the second layer generates the output patterns for classification. All experiments employed fully delayed self-synapses to ensure maximal temporal connectivity. To investigate the effect of network scale, we varied the number of nodes in each layer (16, 32, 64, 128, and 256).

Fig.~\ref{Fig6}(a)–(c) presents the classification accuracy of STD-SNNs (red) and TDA-SNNs (blue) across different node numbers. TDA-SNN accuracy increases steadily with network size, from 92.37$\pm$0.24\% to 98.23$\pm$0.09\% on MNIST, from 85.23$\pm$0.27\% to 89.16$\pm$0.12\% on fMNIST, and from 53.99$\pm$0.86\% to 72.95$\pm$2.02\% on DVS Gesture. At small scales (e.g., 16 nodes), STD-SNN slightly outperforms TDA-SNN, suggesting that limited representational capacity reduces the benefit of temporal multiplexing. As the number of nodes increases, the gap narrows substantially, indicating that TDA-SNN can serve as a competitive feedforward alternative when sufficient temporal nodes are available. These results support the effectiveness of the proposed temporal construction in medium- and large-scale MLP settings, while also showing that its benefits are less pronounced in low-capacity regimes.

%%%%%%%%%%%%%%%%%%%%%%%%%%%%%%%%%%%%%%%%%%%%%%%%Figure7
\begin{figure}[t]
	\centering
	\includegraphics{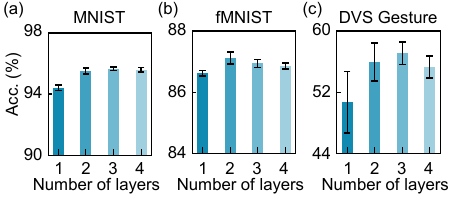}
	\caption{Experimental results of networks with different layer depths on three benchmark datasets. (a) MNIST, (b) fMNIST, and (c) DVS Gesture.}
	\label{Fig7}
\end{figure}
%%%%%%%%%%%%%%%%%%%%%%%%%%%%%%%%%%%%%%%%%%%%%%%%

%%%%%%%%%%%%%%%%%%%%%%%%%%%%%%%%%%%%%%%%%%%%%%%%TEffect of Strategy Selection
\subsubsection{Effect of Strategy Selection}
This subsection investigates the impact of autapse delay selection strategies on TDA-SNN performance in MLP architectures. We compare RD, MC, and time-invariant (T Inv.) strategies across varying numbers of delayed autapses, where the T Inv. strategy assigns identical weights to autapses with the same delay. These experiments evaluate how the choice of delay strategy and weight sharing influences temporal encoding and feature extraction.

Increasing the number of autapses improves classification accuracy across MNIST, fMNIST, and DVS Gesture, as shown in Tab.~\ref{tab:rc_and_mlp}. On MNIST and fMNIST, the three strategies converge quickly once two or more autapses are used, with MC showing a clear disadvantage only in the extreme low-delay regime. On the more temporally demanding DVS Gesture dataset, RD and T-Inv. outperform MC at very small autapse counts, indicating that greater delay diversity is beneficial when temporal structure is complex. As the number of autapses increases beyond four, the gap between strategies narrows and MC becomes comparable to the other two strategies. This ablation suggests that the number of delayed connections is the dominant factor, while the specific selection strategy mainly affects low-resource settings. Statistical significance analyses of the experimental results are provided in Supplementary Sec.~\ref{sec11}. 

%%%%%%%%%%%%%%  tab:Computational Complexity and Memory
\begin{table*}[t]
	\centering
	\caption{Per-layer computational complexity, parameter count, neuron count, speed, and per-neuron information content across different models.}
	\label{tab:CCM}
	\begin{tabular}{@{}llccccc@{}}
		\toprule
		\multirow{2}{*}{Model}   & \multirow{2}{*}{Arch.} &                  \multirow{2}{*}{FLOPs}                   &               \multirow{2}{*}{Params.}                & \multirow{2}{*}{$N_\text{Neu.}$} &   Speed   & $\mathbf{S}$ \\
		                         &                        &                                                           &                                                       &                                  & (s/epoch) &    (bit)     \\ \midrule
		\multirow{3}{*}{STD-SNN}     & RC                     &                          $N^2T$                           &                         $N^2$                         &               $N$                &   22.57   &    489.91    \\
		                         & MLP                    &              $N_{\text{in}}N_{\text{out}}T$               &             $N_{\text{in}}N_{\text{out}}$             &         $N_{\text{out}}$         &   8.36    &   3113.89    \\
		                         & Conv.                  &           $2C_{\text{in}}C_{\text{out}}k^2WHT$            &           $C_{\text{in}}C_{\text{out}}k^2$            &           $C_{out}WH$            &   17.21   &     5.48     \\ \midrule
		\multirow{3}{*}{TDA-SNN} & RC                     &             $\sum_{d} \left ( N-d \right )T$              &            $\sum_{d} \left ( N-d \right )$            &               $1$                &   21.14   &   32096.37   \\
		                         & MLP                    &                      $\sum_{d} N_dT$                      &                    $\sum_{d} N_d$                     &               $1$                &   90.91   &  199237.63   \\
		                         & Conv.                  & $2C_{\text{in}}C_{\text{out}}\left (\sum_dN_d\right )WHT$ & $C_{\text{in}}C_{\text{out}}\left (\sum_dN_d\right )$ &               $1$                &  3060.23  &   56486.07   \\ \bottomrule
	\end{tabular}
\end{table*}

%%%%%%%%%%%%%%%%%%%%%%%%%%%%%%%%%%%%%%%%%%%%%%%%Impact of Layer Number in MLP-Equivalent Structures
\subsubsection{Impact of Layer Number in MLP Structures}
We investigated the effect of network depth ($N_\text{layer}$) on TDA-SNN performance in MLP structures, fixing 64 nodes per layer and employing the RD selection strategy. Classification performance was evaluated across networks with one to four layers. The results are presented in Fig.~\ref{Fig7}.

On MNIST, accuracy steadily increases from 94.41 $\pm$ 0.17\% to 95.60 $\pm$ 0.15\% as depth grows from 1 to 4 layers, indicating that additional depth remains beneficial in this setting. For fMNIST, accuracy rises from 86.62 $\pm$ 0.09\% to 87.12 $\pm$ 0.19\% at 1--2 layers but then decreases slightly at 3 and 4 layers, suggesting that moderate depth is sufficient. On DVS Gesture, performance improves from 1 to 2 layers and then largely saturates, reaching its best value of 57.12 $\pm$ 1.48\% at 3 layers. Thus, the effect of depth is dataset-dependent: deeper networks help on MNIST, moderate depth is adequate on fMNIST, and the benefit saturates early on DVS Gesture. Temporal unfolding supports deeper feedforward constructions, but the optimal depth depends on task complexity and data characteristics. 

Further evidence for the robustness of the MLP setting is provided in Supplementary Sec.~\ref{sec13}. The MNIST learning-rate results confirm stable optimization, and the CIFAR100 results with different numbers of nodes show that TDA-SNN remains competitive on a more challenging dataset.

%%%%%%%%%%%%%%%%%%%%%%%%%%%%%%%%%%%%%%%%%%%%%%%%Figure8
\begin{figure}[t]
	\centering
	\includegraphics{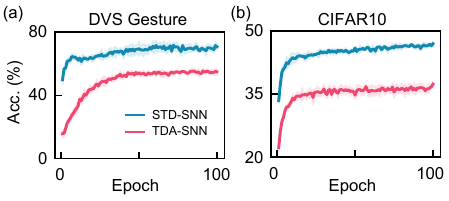}
	\caption{Test accuracy comparison between the TDA-SNN and the STD-SNN under convolutional structure on two benchmark datasets. (a) DVS Gesture and (b) CIFAR-10.}
	\label{Fig8}
\end{figure}
%%%%%%%%%%%%%%%%%%%%%%%%%%%%%%%%%%%%%%%%%%%%%%%%
%%%%%%%%%%%%%%%%%%%%%%%%%%%%%%%%%%%%%%%%%%%%%%%%Convolutional Equivalence Exploration
\subsection{Convolutional Equivalence Exploration}
To examine whether the proposed temporal construction can extend beyond RC and MLP settings, we performed preliminary experiments with convolution-like architectures on DVS Gesture and CIFAR-10, which represent dynamic and static visual inputs, respectively. These experiments are intended as an initial proof of concept for applying TDA within convolutional layers. Further implementation details are provided in Supplementary Sec.~\ref{sec14}.

%The network adopted a standard two-layer convolutional backbone, consisting of a convolutional layer (8 output channels, 7×7 kernel size, stride 1), followed by 4×4 max-pooling and flattening, and a fully connected layer with an output feature size of 512 for classification. The TDA-SNN variant replaced standard synapses in convolutional layers with time-delayed autapses configured using the all delay autapses. All models were trained for 100 epochs using Adam optimizer with a learning rate of 2e-3 and batch size of 64.

Fig.~\ref{Fig8} shows classification performance across training epochs. On DVS Gesture, STD-SNN reaches 73.61 $\pm$ 1.93\% accuracy and stabilizes after approximately 60 epochs, whereas TDA-SNN converges more slowly and plateaus at 57.58 $\pm$ 1.00\%. A similar trend is observed on CIFAR-10, where STD-SNN reaches 47.31 $\pm$ 0.24\% and TDA-SNN stabilizes at 37.64 $\pm$ 0.65\%. These results suggest that, although delayed self-feedback enriches temporal dynamics, it also introduces substantial optimization difficulty in convolutional settings, likely because delayed feedback interferes with hierarchical spatiotemporal feature aggregation. Therefore, the current single-neuron convolution-like construction should be viewed as a preliminary exploration rather than a fully competitive convolutional alternative. Improving this setting will likely require multi-neuron parallelization to better balance spatial representation and temporal recursion.
         
%%%%%%%%%%%%%%%%%%%%%%%%%%%%%%%%%%%%%%%%%%%%%%%%Computational Complexity Analysis
\subsection{Computational Complexity Analysis}
To evaluate efficiency, we analyzed the per-layer complexity of both STD-SNNs and TDA-SNN under three representative architectures: RC, MLP, and convolutional networks. For the RC structure, we used SHD; for MLP, MNIST; and for convolutional networks, CIFAR-10. Both the RC and MLP structures were scaled to a size of 64. We report FLOPs, parameter count, neuron count, and training speed (seconds per epoch). In all cases, TDA-SNN replaces conventional synapses with time-delayed autapses and uses all available autapse delays.

As shown in Table~\ref{tab:CCM}, STD-SNNs exhibit high spatial complexity, with RC and MLP structures scaling quadratically or linearly with the number of connections (\(N^2T\) and \(N_{\text{in}}N_{\text{out}}T\)), whereas TDA-SNN converts these spatial couplings into temporal dependencies (\(\sum_d N_dT\)). This design drastically reduces the number of physical neurons per layer, but the benefit comes at the cost of additional temporal computation due to delayed feedback. The trade-off is mild in the RC setting, where training speed remains comparable, but becomes much more pronounced in MLP and especially convolutional settings. Therefore, the main efficiency advantage of TDA-SNN lies in spatial compactness and neuron-count reduction rather than uniformly lower runtime cost. This observation is consistent with the space--time trade-off discussed throughout the paper.

%%%%%%%%%%%%%%%%%%%%%%%%%%%%%%%%%%%%%%%%%%%%%%%%Discussion on Single-Neuron Storage Capacity
\subsection{Discussion on Single-Neuron Storage Capacity}
We further analyzed the storage capacity of individual neurons by quantifying the per-neuron information content under the same experimental settings as in the previous section for RC, MLP, and convolutional networks. Here, $\mathbf{S}$ denotes the effective information content carried by a single neuron and is computed as $\mathbf{S} = {\text{Acc.} \times N_\text{samples} \times \log_2C}/{N_\text{Neu.}}$. As shown in Table~\ref{tab:CCM}, TDA-SNN achieves substantially higher per-neuron information content than STD-SNN across all three architectures. This increase is mainly due to the extreme reduction in neuron count, from \(N\) or \(N_\text{out}\) in STD-SNN to 1 in TDA-SNN, while preserving useful predictive performance. In the RC and MLP settings, delayed autapses and temporal integration allow a single neuron to encode information that would otherwise be distributed across an entire layer, leading to a dramatic increase in bits per neuron. In the convolution-like setting, the gain is smaller than in RC and MLP because performance is currently more limited, but it still remains far above that of STD-SNN. These results highlight temporal multiplexing as an effective mechanism for increasing information density at the single-neuron level.

%%%%%%%%%%%%%%%%%%%%%%%%%%%%%%%%%%%%%%%%%%%%%%%%Parallelization and Space-Time Trade-off
\subsection{Parallelization and Space--Time Trade-off}
We further examine whether the latency overhead observed in the extreme single-neuron setting can be alleviated by parallelization. Additional results on CIFAR-10 show that the high temporal overhead is specific to the most aggressive compression regime and that TDA-SNN provides a flexible space--time trade-off when more parallel neurons are introduced. As the number of parallel neurons increases, both training and inference time decrease substantially; relative to STD-SNN on CIFAR-10, the training-time overhead drops from 178$\times$ to 46$\times$, and with 512 parallel neurons the inference time reaches 0.92$\times$ of STD-SNN. These observations indicate that the single-neuron model should be viewed as a limit case for maximizing spatial compactness, while parallel multi-neuron realizations provide more practical operating points for deployment. From a memory perspective, this compact formulation remains appealing: on CIFAR-10, the SNN state memory is reduced from 8\,KB in STD-SNN to 4\,Bytes in the single-neuron TDA-SNN. Additional visualizations and scalability results are provided in Supplementary Sec.~\ref{sec15}.

\section{Discussion and Conclusion}
Our study shows that a single TDA-LIF neuron can realize core SNN architectures, including RC, MLP, and convolution-like layers, highlighting the computational potential of individual neurons. Inspired by temporal self-modulation in cerebellar Purkinje cells, TDA-SNN converts temporal feedback into compact spiking computation. Experiments show competitive performance in RC and MLP settings, while convolutional results and parallelization analysis reveal a clear space--time trade-off. TDA-SNN also reduces neuron count and increases per-neuron information content. However, the current single-neuron convolution-like setting remains limited in efficiency and accuracy. Future work will extend this framework to multi-neuron settings and adaptive delay mechanisms.

\section{Acknowledgements}
This work was supported in part by the National Key Research and Development Program of China under Grant 2023YFF1204200, in part by the Brain Science and Brain-like Intelligence Technology-National Science and Technology Major Project under Grant 2022ZD0208500, in part by the Sichuan Science and Technology Program under Grants 2024NSFTD0032, 2024NSFJQ0004, and DQ202410, in part by the Natural Science Foundation of Chongqing, China, under Grant CSTB2024NSCQ-MSX0627, in part by the Science and Technology Research Program of Chongqing Education Commission of China under Grant KJZD-K202401603, and in part by the China Postdoctoral Science Foundation under Grant 2024M763876.

{
    \small
    \bibliographystyle{ieeenat_fullname}
    \bibliography{main}
}

\clearpage
\setcounter{page}{1}
\maketitlesupplementary

\section{Weight Matrix of Autapses}
\label{sec7}
\subsection{Reservoir Computing Model with TDA-LIF}
\begin{figure}[t]
	\centering
	\includegraphics{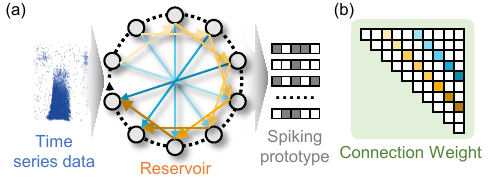}
	\caption{RC architecture of the TDA-SNN model. (a) Overall RC architecture constructed from the TDA-SNN model. (b) Autaptic weight matrix corresponding to the RC framework.}
	\label{FigS1}
\end{figure}

Based on the equivalent formulation introduced in Sec.~3.2, external inputs are injected into each internal temporal node, and the spike activities of all nodes are recorded over the external data time window $T$~\cite{peng2025one, stelzer2021emulating}. The resulting firing patterns are then compared for recognition, as illustrated in Fig.~\ref{FigS1}(a). The autaptic weight matrix $W_a$ characterizes the temporal feedback structure among the unfolded temporal nodes, where each element specifies an autaptic connection from one node to another. Owing to the directed nature of temporal feedback, $W_a$ takes an upper-triangular form, with each parallel diagonal corresponding to a specific autaptic delay $d$. As shown in Fig.~\ref{FigS1}(b), two such diagonals appear, representing delays of $d=3$ (yellow) and $d=6$ (blue). The resulting unfolded structure, derived from and functionally equivalent to the TDA-LIF neuron, forms the complete reservoir-computing (RC) architecture.
%%%%%%%%%%%%%%%%%%%%%%%%%%%%%%%%%%%%%%%%%%%%%%%%%Figure1
%%%%%%%%%%%%%%%%%%%%%%%%%%%%%%%%%%%%%%%%%%%%%%%%%

\subsection{Multilayer Perceptron Model with TDA-LIF}
\begin{figure}[t]
	\centering
	\includegraphics{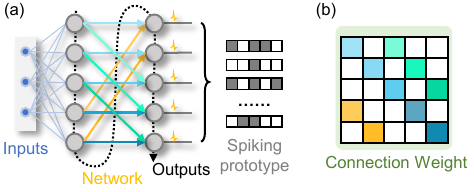}
	\caption{MLP architecture of the TDA-SNN model. (a) Overall MLP architecture constructed from the TDA-SNN model. (b) Autaptic weight matrix corresponding to the MLP framework.}
	\label{FigS2}
\end{figure}

Similarly, an external layer is added before the first TDA-SNN layer to form a multilayer perceptron (MLP), and the spike activity of the final layer is recorded over the external data time window $T$~\cite{stelzer2021deep}. The resulting spike patterns are then compared for recognition, as shown in Fig.~\ref{FigS2}(a). The autaptic connection matrix $W_a$ describes the temporal feedback structure across temporal nodes: each element specifies an autaptic connection from a node in the previous layer to one in the next layer, and each parallel diagonal corresponds to a particular autaptic delay. As illustrated in Fig.~\ref{FigS2}(b), three diagonals appear, representing delays of $d=2$ (yellow), $d=5$ (blue), and $d=7$ (green). This delayed-feedback structure unfolds directly from the TDA-LIF neuron and is functionally equivalent to it, thereby forming the complete spiking MLP architecture.
%%%%%%%%%%%%%%%%%%%%%%%%%%%%%%%%%%%%%%%%%%%%%%%%%Figure2
%%%%%%%%%%%%%%%%%%%%%%%%%%%%%%%%%%%%%%%%%%%%%%%%%

\section{Spatiotemporal Prototype Learning}
\label{sec8}
To enable class-specific spatiotemporal representations within TDA-SNN, we construct a set of binary prototypes corresponding to the desired spiking patterns of each class. These prototypes serve as reference patterns against which the encoded network outputs are compared, facilitating classification and similarity measurement in the spatiotemporal domain.

Specifically, a set of binary prototypes $\mathbf{K} \in \{0, 1\}^{C \times T}$ is defined to represent target spiking patterns for $C$ classes over $T$ timesteps. Let $f(x;\theta)$ denote the encoded spatiotemporal representation of input $x$. To measure the similarity between the encoded representation and each class prototype, we compute
\begin{equation} 
	d_i(x) = -\left\| f(x;\theta) - \mathbf{k}_i \right\|_2^2, \label{eq6}
\end{equation}
where $d_i(x)$ denotes the similarity score between $f(x;\theta)$ and the $i$-th prototype $\mathbf{k}_i \in \mathbf{K}$.

During inference, each input sample is assigned to the class of the closest prototype in the spatiotemporal feature space. The prototype-based classification procedure follows the method described in~\cite{cai2025robust, wu2018spatio, wu2019direct}.

%%%%%%%%%%%%%%  tab:hyperparam
\begin{table}
	\caption{The hyperparameters for different datasets.}
	\label{tab:hyperparam}
	\centering
	\begin{tabular}{@{}lccccccc@{}}
		\toprule
		Dataset     & bs  &  lr   & $T$ & $\text{d}t$ & $v_{\text{th}}$ & $\tau$ &  \\ \midrule
		DEAP        & 200 & 0.002 &  6  &     100     &       1.0       &  0.5   &  \\
		SHD         & 512 & 0.01  & 100 &     14      &       0.3       &  0.3   &  \\
		MNIST       & 512 & 0.002 &  4  &      -      &       0.3       &  0.5   &  \\
		fMNIST      & 512 & 0.002 &  4  &      -      &       0.3       &  0.5   &  \\
		DVS Gesture & 128 & 0.002 &  8  &     125     &       1.1       &  0.67  &  \\
		CIFAR10     & 200 & 0.002 &  4  &      -      &       1.0       &  0.5   &  \\ 
		CIFAR100    & 512 & 0.002 &  4  &      -      &       1.0       &  0.5   &  \\\bottomrule
	\end{tabular}
\end{table}

\section{Datasets and Experimental Setup}
\label{sec9}
We evaluate TDA-SNN across multiple benchmarks under different structural mappings. DEAP~\cite{koelstra2011deap} and SHD~\cite{cramer2020heidelberg} are used to evaluate the reservoir-computing setting; MNIST~\cite{deng2012mnist}, Fashion-MNIST (fMNIST)~\cite{xiao2017fashion}, and DVS Gesture~\cite{amir2017low} are used to validate the MLP setting; and DVS Gesture together with CIFAR10~\cite{krizhevsky2009learning} are further used to study the convolution-like setting.

All models are implemented in PyTorch and trained on an NVIDIA A100 GPU using the Adam optimizer with a cosine learning-rate decay schedule. Unless otherwise stated, training is conducted for 100 epochs. For RC and MLP experiments, results are averaged over 10 independent runs and reported as mean $\pm$ standard deviation; due to the higher computational cost of convolutional experiments, those results are averaged over 5 runs. To ensure a fair comparison, TDA-SNN and the corresponding standard SNN baselines use aligned training protocols and matched architectural scales for each setting.

The complete set of hyperparameters is summarized in Table~\ref{tab:hyperparam}, including batch size (bs), learning rate (lr), time window $T$, slice width $\mathrm{d}t$, membrane threshold $v_{\text{th}}$, and membrane decay factor $\tau$.

In our experiments, the standard SNN evolves over an external time window, whereas TDA-SNN introduces an additional internal evolution time. Consequently, the membrane potential dynamics in TDA-SNN are decoupled from those of the standard SNN: the latter progresses along external timesteps, while the former evolves over internal temporal nodes. In TDA-SNN, the external time window is treated as an independent dimension and does not directly participate in the neuron dynamics.

%%%%%%%%%%%%%%%%%%%%%%%%%%%%%%%%%%%%%%%%%%%%%%%%%Figure3
\begin{figure}[t]
	\centering
	\includegraphics{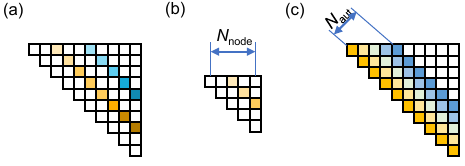}
	\caption{Evolution of connection weights in the RC architecture. (a) Random delay strategy. (b) Low-node configuration. (c) Maximum connection strategy.}
	\label{FigS3}
\end{figure}
%%%%%%%%%%%%%%%%%%%%%%%%%%%%%%%%%%%%%%%%%%%%%%%%%

%%%%%%%%%%%%%%%%%%%%%%%%%%%%%%%%%%%%%%%%%%%%%%%%%Figure4
\begin{figure}[t]
	\centering
	\includegraphics{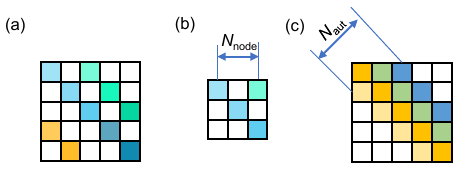}
	\caption{Evolution of connection weights in the MLP architecture. (a) Random delay strategy. (b) Low-node configuration. (c) Maximum connection strategy.}
	\label{FigS4}
\end{figure}
%%%%%%%%%%%%%%%%%%%%%%%%%%%%%%%%%%%%%%%%%%%%%%%%%

\section{Weight Matrices Under RC and MLP Architectures}
\label{sec10}
In the RC architecture, the TDA-SNN connection matrix is an upper-triangular square matrix. As shown in Fig.~\ref{FigS3}, the autaptic selection strategy and the number of nodes jointly determine its structure. The random delay (RD) strategy produces a sparser matrix with generally longer-delay connections. Increasing the number of nodes expands the space of possible connections, whereas using fewer nodes naturally reduces the total number of available connections. In contrast, selecting autapses along the main diagonal toward the upper-right corner under the maximum-connections (MC) strategy yields the densest connectivity while simultaneously minimizing autaptic latency.

For the MLP architecture, the TDA-SNN connection matrix width and height correspond to the number of input and output nodes, respectively. Fig.~\ref{FigS4} illustrates how autapse selection strategies shape the matrix. Using the MC strategy, autapses are selected along the diagonal and distributed to neighboring nodes, maximizing connectivity while minimizing latency. In contrast, the RD strategy yields sparser, longer-delayed connections. This demonstrates that, in feedforward structures, structured delay assignment enables efficient temporal propagation across layers, analogous to the role of weight connectivity in conventional MLPs.

%%%%%%%%%%%%%  tab:rc_and_mlp
\begin{table*}[t]
	\renewcommand{\arraystretch}{1.0}
	\caption{Paired t-tests of Autaptic Selection Strategies Across Different Delays. * denotes $p<0.05$.}
	\label{tab:p_value_rc_and_mlp}
	\centering
	\setlength{\tabcolsep}{5.0mm}{
		\begin{tabular}{@{}ccccccccc@{}}
			\toprule
			Dataset          &   Delay   &  1   &  2   &  4   &  8   &  16  &  32  &  64  \\ \midrule
			DEAP      &   MC-RD   & 0.79 & 0.66 & 0.65 & 0.73 & 0.51 & 0.05 & 1.0  \\ \midrule
			SHD      &   MC-RD   & 0.53 &  *   &  *   &  *   & 0.40 & 0.81 & 1.0  \\ \midrule
			\multirow{3}*{MNIST}    &   MC-RD   &  *   &  *   &  *   &  *   &  *   &  *   &  *   \\
			& MC-T Inv. &  *   & 0.06 &  *   & 0.65 & 0.97 & 0.95 &  *   \\
			& RD-T Inv. & 0.35 & 0.88 &  *   &  *   &  *   &  *   &  *   \\ \midrule
			\multirow{3}*{fMNIST}  &   MC-RD   &  *   &  *   & 0.59 &  *   &  *   &  *   &  *   \\ 		                       
			& MC-T Inv. &  *   &  *   &  *   & 0.54 & 0.60 & 0.53 &  *   \\		                       
			& RD-T Inv. & 0.53 & 0.68 & 0.06 &  *   &  *   &  *   &  *   \\ \midrule
			\multirow{2}*{DVS}  &   MC-RD   &  *   & 0.58 & 0.30 & 0.06 &  *   &  *   &  *   \\ 		
			\multirow{2}*{Gesture} & MC-T Inv. &  *   & 0.26 & 0.08 & 0.17 & 0.92 & 0.48 &  *   \\		                       
			& RD-T Inv. & 0.26 & 0.57 & 0.44 & 0.72 &  *   &  *   &  *   \\ \midrule
	\end{tabular}}
\end{table*}

\section{Significance Analysis of Experimental Results}
\label{sec11}
The statistical significance results are summarized in Table~\ref{tab:p_value_rc_and_mlp}. Overall, the effect of autaptic selection strategy depends on both the dataset and the delay budget. On DEAP, most comparisons are not statistically significant, suggesting that the task is relatively insensitive to the specific delay-selection strategy. On SHD, significant differences appear mainly at intermediate delays, indicating that strategy choice matters in specific temporal regimes. On MNIST, many pairwise comparisons are significant, especially for MC-RD and for RD-T Inv. at moderate-to-large delays, showing that strategy selection can substantially affect performance. fMNIST and DVS Gesture also exhibit several significant differences, although the pattern varies with both the comparison pair and the delay value rather than following a uniform trend. These observations are consistent with the ablation results in showing that the impact of delay assignment is dataset-dependent and becomes particularly evident in specific delay regimes.

%%%%%%%%%%%%%%%%%%%%%%%%%%%%%%%%%%%%%%%%%%%%%%%%%Figure5
\begin{figure}[t]
	\centering
	\includegraphics{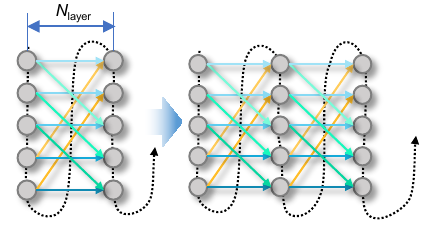}
	\caption{Extending TDA-SNN to multi-layer deep models under the MLP architecture.}
	\label{FigS5}
\end{figure}
%%%%%%%%%%%%%%%%%%%%%%%%%%%%%%%%%%%%%%%%%%%%%%%%%

\section{Impact of Layer Number in MLP Structures}
\label{sec12}
TDA-SNN can be extended to deep architectures by increasing the number of internal nodes, allowing a single TDA-LIF neuron to emulate multiple layers through its temporal unfolding. As shown in Fig.~\ref{FigS5}, connections with the same color denote autapses sharing the same delay value, which remain consistent across layers. This preserves the structured temporal feedback while enabling the construction of deeper computational models.

\section{Additional Ablation Studies}
\label{sec13}
We additionally provide two supplementary ablation studies: learning-rate sensitivity on MNIST and an ablation with different numbers of nodes on CIFAR100. These experiments further examine the robustness of TDA-SNN under different optimization and architectural settings.

%%%%%%%%%%%%%%%%%%%%%%%%%%%%%%%%%%%%%%%%%%%%%%%%%Figure6
\begin{figure}[t]
	\centering
	\includegraphics{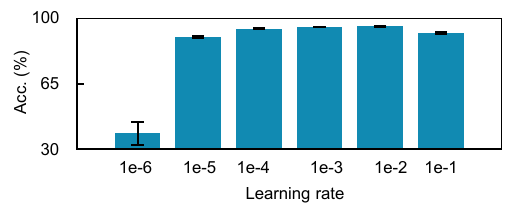}
	\caption{Learning rate sensitivity study for the MLP architecture on MNIST.}
	\label{FigS6}
\end{figure}
%%%%%%%%%%%%%%%%%%%%%%%%%%%%%%%%%%%%%%%%%%%%%%%%%
\subsection{MNIST}
\begin{figure}[t]
	\centering
	\includegraphics{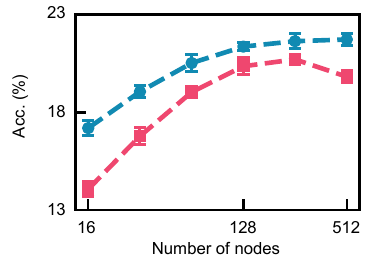}
	\caption{Number of nodes ablation on CIFAR100 for STD-SNN and TDA-SNN.}
	\label{FigS7}
\end{figure}

For MNIST, the learning-rate ablation is conducted using the hyperparameter setting in Table~\ref{tab:hyperparam}, varying the initial learning rate from $10^{-6}$ to $10^{-1}$ while keeping the remaining settings unchanged. The corresponding mean accuracies with standard deviations are visualized in Fig.~\ref{FigS7}. Although the best performance is achieved at a learning rate of $10^{-2}$ (95.86$\pm$0.26\%), we use $10^{-3}$ (95.55$\pm$0.20\%) as the default setting because it remains close to the optimum while providing a more stable optimization process and stronger convergence robustness. This choice is particularly important for SNNs, whose discrete spikes and highly nonlinear dynamics make training sensitive to overly large learning rates, which can lead to oscillation and instability.

%%%%%%%%%%%%%%%%%%%%%%%%%%%%%%%%%%%%%%%%%%%%%%%%%Figure7
%%%%%%%%%%%%%%%%%%%%%%%%%%%%%%%%%%%%%%%%%%%%%%%%%
\subsection{CIFAR100}
\begin{figure}[t]
	\centering
	\includegraphics{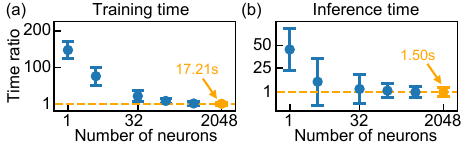}
	\caption{Training (a) and inference (b) time ratios of TDA-SNN relative to STD-SNN on CIFAR10. The yellow marker denotes STD-SNN (ratio = 1.0). The x-axis is logarithmic.}
	\label{FigS8}
\end{figure}

For CIFAR100, we provide an additional ablation with different numbers of nodes under a more challenging visual classification setting. Following the setting of Section 4.3.1, we use a two-layer network, and the hyperparameter configuration is given in Table~\ref{tab:hyperparam}. The number of nodes is varied over $\{16, 32, 64, 128, 256, 512\}$ while the remaining architectural and training settings are kept fixed. As shown in Fig.~\ref{FigS8}, STD-SNN improves steadily from 17.20$\pm$0.39\% at 16 nodes to 21.72$\pm$0.32\% at 512 nodes. TDA-SNN also improves substantially, from 14.08$\pm$0.39\% at 16 nodes to a best accuracy of 20.70$\pm$0.21\% at 256 nodes. When the number of nodes is further increased to 512, the accuracy decreases slightly to 19.82$\pm$0.31\%, suggesting that simply increasing model size does not always yield additional gains on this challenging dataset. Nevertheless, TDA-SNN consistently improves as the number of nodes increases over a broad range and retains strong performance at medium-to-large scales, demonstrating that the proposed model remains effective even in a more difficult setting.

\section{Implementation Details for Convolution-like Settings}
\label{sec14}
The network adopted a standard two-layer convolutional backbone, consisting of a convolutional layer (8 output channels, 7$\times$7 kernel size, stride 1, padding 3), followed by 4$\times$4 max-pooling and flattening, and a fully connected layer with an output feature size of 512 for classification. The TDA-SNN variant replaced standard synapses in convolutional layers with time-delayed autapses configured using all available autaptic delays.

\section{Parallelization and Scalability in Convolution-like Settings}
\label{sec15}
To further analyze the temporal overhead of TDA-SNN, we study how parallelization changes the runtime characteristics in convolution-like settings. The extreme latency reported is specific to the single-neuron setting, which is designed to probe the limit of spatial compression. As shown in Fig.~\ref{FigS6}, increasing the number of parallel neurons reduces both training and inference time. On CIFAR10, relative to STD-SNN, the training-time overhead decreases from 178$\times$ to 46$\times$, and with 512 neurons the inference time reaches 0.92$\times$ of STD-SNN. These results further support that TDA-SNN offers a flexible space--time trade-off rather than a fixed high-latency operating point.

%%%%%%%%%%%%%%%%%%%%%%%%%%%%%%%%%%%%%%%%%%%%%%%%%Figure8
%%%%%%%%%%%%%%%%%%%%%%%%%%%%%%%%%%%%%%%%%%%%%%%%%

We additionally evaluate a multi-neuron extension to examine scalability when the compression constraint is relaxed. Using 512 neurons for DVS Gesture and 256 neurons for CIFAR10 improves accuracy from 57.78\% to 62.43\% on DVS Gesture and from 37.64\% to 40.42\% on CIFAR10. Although a gap to STD-SNN remains, this upward trend confirms that the proposed TDA mechanism scales effectively when moving beyond the extreme single-neuron regime.

%{
%	\small
%	\bibliographystyle{ieeenat_fullname}
%	\bibliography{main_s}
%}
%\end{document}

% WARNING: do not forget to delete the supplementary pages from your submission 
% \input{sec/X_suppl}

\end{document}